\documentclass[letterpaper]{article} % DO NOT CHANGE THIS
\usepackage[preprint]{aaai2027}  % DO NOT CHANGE THIS
\usepackage[hyphens]{url}  % DO NOT CHANGE THIS
\usepackage{graphicx} % DO NOT CHANGE THIS
\usepackage{natbib}  % DO NOT CHANGE THIS AND DO NOT ADD ANY OPTIONS TO IT
\usepackage{caption} % DO NOT CHANGE THIS AND DO NOT ADD ANY OPTIONS TO IT
\usepackage{booktabs}
\usepackage{multirow}
\usepackage{pifont}
\newcommand{\cmark}{\ding{51}}
\newcommand{\xmark}{\ding{55}}
\usepackage[table]{xcolor}
\definecolor{oursblue}{RGB}{235,244,250}

\newcommand{\bench}{\textsc{FoR-T2I}}

\title{Can Text-to-Image Models Draw from the Right Frame of Reference?}
\author{
    Zheyuan Gu\textsuperscript{\rm 1,\rm 2},
    Ruihang Li\textsuperscript{\rm 1},
    Yong Huang\textsuperscript{\rm 1},
    Yiqian Zhang\textsuperscript{\rm 1},
    Xiangzhao Hao\textsuperscript{\rm 1},\\
    Jiaxin Niu\textsuperscript{\rm 1},
    Jiahao Hu\textsuperscript{\rm 1},
    Zhenyu Zhang\textsuperscript{\rm 1}\corresponding
}
\affiliations{
    \textsuperscript{\rm 1}ERNIE Team, Baidu Inc.\\
    \textsuperscript{\rm 2}Peking University
}

\begin{document}

\maketitle

\begin{abstract}
Spatial instruction following has become a crucial requirement for text-to-image (T2I) generation.
A common challenge arises when directional expressions are interpreted under different frames of reference.
For example, ``the left of'' may refer to the viewer's image coordinates or to the intrinsic orientation of an object, leading to different expected layouts.
Existing T2I benchmarks reveal important layout failures, yet they rarely isolate whether models can follow a specified frame of reference when it differs from camera view.
To mitigate this gap, we introduce \textbf{\bench}, a benchmark for evaluating this distinction with 1,200 prompt pairs built from controlled spatial layouts.
In each pair, the camera-view (Cam) prompt states the target relation in camera view, while the frame-of-reference (FoR) prompt describes the same target placement through an oriented anchor object.
Across 22 closed-source and open-source T2I models, mean final accuracy is 41.8\% lower on FoR prompts than on matched Cam prompts; even the best-performing model achieves only 44.3\% FoR accuracy.
This suggests that current models struggle more when the same layout is described through an object's orientation rather than directly in image coordinates.
We further analyze this gap by relation type and camera view, compare several training-free prompting and feedback-based mitigation strategies, and propose a VLM-gated rewriting approach that selects rewritten prompts using visual feedback, improving average FoR accuracy from 25.0\% to 29.2\% under the same generation budget.

\end{abstract}

\section{Introduction}

\begin{figure}
    \centering
    \includegraphics[width=1.0\linewidth]{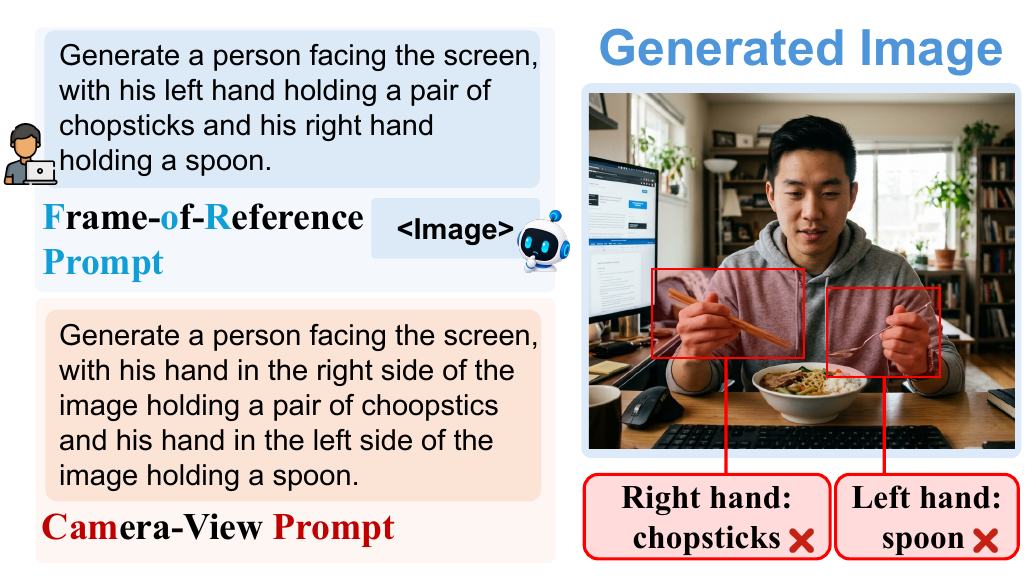}
    \caption{
An example of reference-frame confusion in text-to-image generation.
The generated image follows the image frame and swaps the requested hand-object assignments.
}
    \label{fig:teaser}
    \vspace{-0.2cm}
\end{figure}

% T2I进展，but得根据prompt画对spatial结构，teaser show failure
Recent text-to-image (T2I) models have achieved substantial improvements in visual quality and instruction following \citep{survey1,gan,rombach2022highresolutionimagesynthesislatent,lipman2023flowmatchinggenerativemodeling}. 
For real-world applications such as design assistance, embodied simulation, and visual planning
\citep{zhang2023texttoimage, yang2023diffusionsurvey, cao2024controllable, liu2026diffusionrobotics, ding2025worldmodels},
a generated image is useful only when it preserves the spatial structure specified by the prompt.
This requirement is especially challenging for directional language.
As shown in Figure~\ref{fig:teaser}, a prompt asking for a person \emph{facing the viewer} with \emph{chopsticks in his left hand and a spoon in his right hand} yields a plausible image that nonetheless places the chopsticks on the image's left.

% FoR是一个spatial language的基础概念，一句话讲解，再讲解T2I里这个概念的意义，提及teaser说明这是一个FoR failure，目前的evaluation没有
This error is not merely a failure to render the requested objects. Because the person faces the viewer, his anatomical left hand appears on the right side of the image, and his anatomical right hand appears on the left side. The model instead follows the viewer's image frame when the prompt specifies the person's own left and right hands.
We refer to this failure as reference-frame confusion: the model uses the wrong frame of reference for a directional expression.
Frames of reference are a basic property of spatial language \citep{levinson2003space}; terms such as ``left'', ``right'', ``front'', and ``behind'' must be interpreted with respect to some reference frame, such as the viewer's image frame or an object's own orientation.
This makes reference-frame confusion a generation problem that must be evaluated separately from whether the image is plausible or contains the requested objects.

% xianzai1

% [ZZY] 这里应该先T2I起手，然后说最近有一些工作在研究compositional prompt中object的位置关系，可以弱化spatial的表述
Recent T2I evaluations have increasingly studied whether models can follow compositional prompts involving multiple objects and their relations.
Benchmarks such as VISOR, GenEval, and T2I-CompBench++ evaluate object presence, counting, attribute binding, and object-object relations, while GenSpace and SpatialGenEval further extend evaluation to richer spatial scenes and 3D-aware configurations \citep{gokhale2022benchmarking, ghosh2023geneval, huang2023t2icompbench, wang2025genspace, wang2026spatialgeneval}.
These studies provide important evidence that current models still struggle with object layouts and relational prompts.
% [ZZY] 应该是说他们没有对frames of reference这个问题展开系统性的分析，也难以度量模型在给定参考系下的效果差异，相当于我们一方面是引入了frames of reference的概念，另一方面是给了系统性的分析（pair-wise prompts）
However, they do not systematically analyze the role of frames of reference in such prompts.
In particular, they usually do not compare two prompts that describe the same target layout under different frames of reference.
This makes it difficult to tell whether a model is truly following the specified frame of reference or simply defaulting to the camera view.
We therefore ask:
\textbf{can T2I models draw the correct layout when the specified frame of reference diverges from the camera view?}

To answer this question, we introduce \textbf{\bench}, a benchmark for frame-of-reference-aware  generation in T2I models.
Each test case specifies a target spatial layout and instantiates it as a pair of camera-view (Cam) and FoR prompts. 
The \textbf{Cam prompt} describes the layout using direction words from the camera view.
The \textbf{FoR prompt} describes the same layout in the anchor's intrinsic frame, so that satisfying it generally requires placing the target at a \emph{different} camera-view position.
Because each pair shares the same objects, scene context, and target layout, their performance difference provides an estimate of the additional difficulty associated with FoR descriptions.
\bench~proposes three task levels: Level 1 tests a single FoR transformation, Level 2 combines multiple relations under a shared anchor, and Level 3 requires chained orientation before spatial transformation.
% [ZZY] 这里的第三个level还是不够直观，底下讲的挺直观的，看看是不是能在图1里标一下什么是anchor，什么是target
% \bench  contains three difficulty levels, from single-anchor relations to multi-relation scenes and multi-step orientation inference.

We evaluate 22 recent closed-source and open-source T2I models on 1,200 prompt pairs.
To identify where FoR generation fails, we separately assess whether models place target objects correctly and render anchors with the required orientations, and analyze how performance varies across camera viewpoints and reasoning levels.
Every evaluated model achieves lower accuracy on FoR prompts than on its paired Cam prompts.
Averaged across the eight closed-source models, final accuracy drops from 50.6\% on Cam prompts to 29.1\% on FoR prompts, a decrease of 21.5 percentage points.
We further evaluate several training-free prompting mitigations, and propose a VLM-gated rewriting strategy that selects among rewritten prompts using visual feedback, improving average FoR accuracy from 25.0\% to 29.2\% under a matched generation budget.
% 数值，写的稍微多一点，扩写，给出一些结论，复现了几个baselinebalabala，效果不好

Our contributions are threefold.
\begin{itemize}
    \item We formulate frame-of-reference grounding in T2I generation as an evaluation problem and introduce \bench, containing 1200 matched Cam--FoR prompt pairs across three task levels.
    \item We benchmark 22 recent T2I models and find that mean final accuracy is 41.8\% lower on FoR prompts than on matched Cam prompts, while even the best model achieves only 44.3\% FoR accuracy.
    \item We introduce a training-free, VLM-gated prompt rewriting strategy that uses visual feedback to improve average FoR accuracy from 25.0\% to 29.2\% with the same budget.
\end{itemize}

\section{Related Work}
% \noindent \textbf{Benchmarking Text-to-Image Models.}
% Existing T2I benchmarks evaluate complementary aspects of instruction following. 
% Composition-oriented benchmarks, such as GenEval and T2I-CompBench++, assess object presence, attribute binding, counting, and relational composition \citep{ghosh2023geneval, huang2023t2icompbench}.
% Broader instruction-following benchmarks, including TIIF-Bench and OneIG-Bench, extend evaluation to style, layout, text rendering, and complex prompts \citep{titf, oneig}.
% % [ZZY] 引用挂了
% More directly relevant to our work are benchmarks for spatial generation. VISOR evaluates object--object spatial relations \citep{gokhale2022benchmarking}, while GenSpace and SpatialGenEval extend evaluation to richer 3D-aware spatial configurations \citep{wang2025genspace, wang2026spatialgeneval}.
% Together, these benchmarks have expanded T2I evaluation from isolated object relations to compositional layouts and more complex spatial scenes.
\noindent\textbf{Frames of Reference in Spatial Tasks.}
Spatial descriptions can be grounded in intrinsic, relative, or absolute frames of reference \citep{levinson2003space, taylor1996perspective,tenbrink2011reference,carlsonradvansky1997reference}.
\footnote{Following this taxonomy, camera view corresponds to the relative frame. For brevity, we write Cam for camera-view prompts and reserve FoR for prompts requiring the anchor's intrinsic frame; we do not consider the absolute frame, which is under-determined in a single image.}
For example, ``left'' may refer to the viewer's left or to an oriented object's own left.
FoREST evaluates frame-dependent spatial descriptions in language models \citep{premsri2025forest}, while ViewSpatial-Bench and allocentric perception benchmarks study perspective-dependent interpretation in vision-language models \citep{li2025viewspatial, wang2026allocentric}.
These works establish the reference frame as an important factor in spatial interpretation.
Their focus, however, is on interpreting language or pre-existing visual observations, where the anchor's orientation is given. In generation it's produced by the model, so frame resolution and orientation rendering must be separated by the evaluation protocol.

\begin{figure*}
    \centering
    \includegraphics[width=0.9\textwidth]{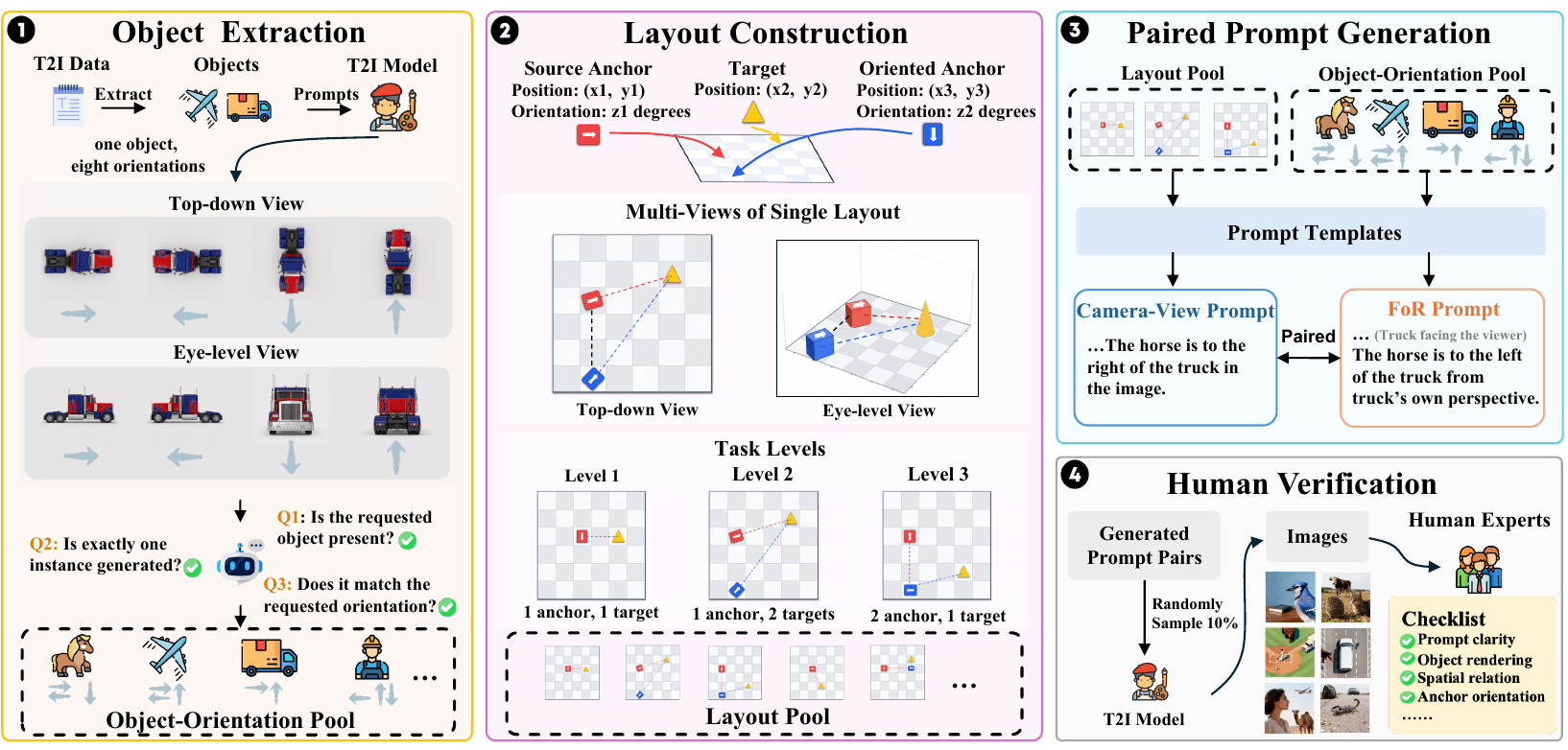}
    \caption{
\textbf{\bench\ construction pipeline.}
\textbf{(1)} We extract objects with a well-defined intrinsic front and keep 
renderings that pass presence, single-instance, and orientation checks.
\textbf{(2)} Each layout deterministically specifies the anchor position, 
target position and anchor orientation.
\textbf{(3)} Each layout yields a Camera-View and a FoR prompt that differ only in whether the spatial term is anchored to the camera or to the explicitly named anchor's intrinsic frame.
\textbf{(4)} $10\%$ of pairs are audited by human experts, full details are reported in the appendix.
}
    \label{fig:benchmark_construction}
\end{figure*}

\begin{table*}[t]
\centering
\small
\renewcommand{\arraystretch}{1.08}
\begin{tabular*}{\textwidth}{
@{\extracolsep{\fill}}llccc@{}}
\toprule
Benchmark
& Scale
& Object-Relative
& FoR Contrast
& Multi-step FoR \\
\midrule

VISOR~\citep{gokhale2022benchmarking}
& 25,280 prompts
& \xmark & \xmark & \xmark \\

GenEval~\citep{ghosh2023geneval}
& 553 prompts
& \xmark & \xmark & \xmark \\

T2I-CompBench++~\citep{huang2023t2icompbench}
& 6,000 prompts
& \xmark & \xmark & \xmark \\

GenSpace~\citep{wang2025genspace}
& 1,800 prompts
& \cmark & \xmark & \xmark \\

SpatialGenEval~\citep{wang2026spatialgeneval}
& 1,230 prompts
& \cmark & \xmark & \xmark \\

\textbf{\bench}
& \textbf{1,200 prompt pairs}
& \cmark & \cmark & \cmark \\
\bottomrule
\end{tabular*}

\caption{
Comparison with spatial text-to-image benchmarks.
Object-Relative indicates directions defined from a reference object's perspective.
FoR Contrast indicates a controlled comparison between object-relative and viewer-centered descriptions. Multi-step FoR indicates tasks in which one FoR relation must be resolved
before another.
}
\label{tab:benchmark_comparison}
\end{table*}

\noindent\textbf{Benchmarking Text-to-Image Models.}
Existing T2I benchmarks evaluate complementary aspects of instruction
following.
Composition-oriented benchmarks, such as GenEval and T2I-CompBench++, assess
object presence, attribute binding, counting, and relational composition
\citep{ghosh2023geneval, huang2023t2icompbench}.
Broader benchmarks, including TIIF-Bench and OneIG-Bench, cover style, layout,
text rendering, and complex prompts \citep{titf, oneig}.
Spatially focused benchmarks evaluate generated object relations and scene
layouts.
VISOR studies object--object relations \citep{gokhale2022benchmarking}, while
GenSpace and SpatialGenEval cover richer spatial configurations and 3D-aware
relations \citep{wang2025genspace, wang2026spatialgeneval}.
These benchmarks determine whether a generated image satisfies a requested layout, but do not isolate the effect of expressing that layout in viewer- or object-centered terms.
The impact of reference-frame choice on T2I generation remains unmeasured.

\section{\bench\ Benchmark}
\subsection{Task Levels}
% [ZZY] 得讲一下啥是anchor，啥是target，感觉还是有图直观些
Each item in \bench\ starts from a concrete target layout with fixed objects, camera view, anchor orientation, and ground-truth spatial placement.
Here, an \textbf{anchor} is the oriented object that provides the reference frame, and a \textbf{target} is the object that the prompt places relative to the anchor.
We verbalize each layout as a paired Cam and FoR prompt.
The Cam prompt describes the target placement directly in the camera view, while the FoR prompt describes the same placement relative to the anchor's own left, right, front, or back.
Thus, the two prompts share the same expected layout but differ in whether the model can place the target directly from the camera view or must use the anchor's facing direction.

\begin{itemize}{
\item \textbf{Level 1: Single FoR.}
Each scene contains one anchor and one target. The FoR prompt places the target relative to the anchor's own left, right, front, or back; the paired Cam prompt places it in the viewer's image frame. This level requires one frame conversion: mapping the anchor-relative relation to the correct camera-view position.

% \item \textbf{Level 2: Mixed reference frames.}
% Each scene contains one anchor and two targets.
% One target is placed relative to the anchor's own left, right, front, or back, while the other is placed directly in the viewer's image frame.
% This level requires the model to handle an anchor-relative relation and an image-frame relation within the same scene.

\item \textbf{Level 2: Mixed FoR.} Each scene contains one anchor and two targets. The FoR prompt places one target relative to the anchor's own left, right, front, or back, and the other in the viewer's image frame; the paired Cam prompt places both in the image frame. This level requires resolving two directional expressions under different frames.

\item \textbf{Level 3: Chained FoR.} Each scene contains two anchors and one target. 
The FoR prompt specifies where the second anchor is placed relative to the first anchor, and then specifies the second anchor's facing direction, with both relations defined in the first anchor's frame of reference.
This level chains two frame conversions: the second anchor's orientation must be resolved before placing the target.}
\end{itemize}

\subsection{Benchmark Construction}
% 我们用一个layout合成引擎，合成了top-down视角 & eye-level视角下的一些空间layout，
% maybe 有审稿人会问：为什么不做prompt rewrite，为什么不用更加自由的开放式生成来做prompt generation而使用template
% balabala，这里最后给出1，200的数字，然后compare with other benchmark

Figure~\ref{fig:benchmark_construction} summarizes the benchmark construction pipeline.
We first construct an object vocabulary from prompts in existing T2I benchmarks \citep{ghosh2023geneval, huang2023t2icompbench, wang2026spatialgeneval}.
An off-the-shelf linguistic parser extracts noun phrases, which are then normalized and deduplicated.
We manually retain concrete, visually recognizable objects whose front--back axis is clear enough for their left and right sides to be consistently defined.

We next build an object-orientation pool to avoid using object-view-orientation configurations whose facing direction cannot be reliably judged.
Using single-object prompts that specify only the camera view and the object's orientation, we generate images for each retained object under two camera views (eye-level and top-down) and four canonical orientations per view, yielding eight candidate configurations.
A VLM-based judge checks object presence, instance count, and orientation consistency with the prompt.
Only the object-view-orientation configurations that pass all checks are kept for later prompt instantiation.

Independently of object identity, we procedurally construct abstract layouts in a discrete spatial coordinate system.
Each layout specifies the entity slots required by its task level, their coordinates, the camera view, and the orientation of each anchor.
We discard layouts in which entities overlap, fall outside the valid spatial region.
For each accepted layout, the engine derives two relation labels from the same coordinates: a camera-view relation from the projected image positions, and a FoR relation from the target displacement relative to the oriented anchor.
We then select layouts to obtain approximately balanced coverage across task levels, camera views, anchor orientations, and camera-view relations.

Finally, we instantiate each selected layout with concrete object categories from the Object-Orientation Pool.
For every orientation-constrained anchor slot, we sample an object whose corresponding camera-view and orientation configuration is available.
Anchor and target categories are kept distinct, and frequency-aware sampling prevents a small number of common objects from dominating the benchmark.

Each instantiated layout is verbalized using matched prompt templates: the Cam prompt states the camera-view relation, while the FoR prompt states the corresponding relation relative to the oriented anchor. 
We use templated verbalization rather than free-form generation because the two prompts in a pair must differ only in the frame in which the direction word is resolved; free-form paraphrasing would introduce uncontrolled lexical and syntactic variation, confounding the comparison the benchmark is designed to isolate. 
This process produces 1,200 Cam--FoR prompt pairs across the three task levels. Appendix reports our benchmark's statistics.

\section{Training-Free FoR Mitigation}
\label{sec:training_free}

% [ZZY] 开头需要一个动机，比如说现在prompt rewrite是能够提升文生图效果的一个主要方法，我们也尝试用这个来解决FoR的问题
Prompt rewriting is the most widely deployed way to improve instruction following in T2I systems, and is a natural first attempt for FoR prompts: if a rewrite makes the implied camera-view placement explicit, the model no longer performs the frame conversion itself.
But a rewrite may resolve the frame incorrectly, or alter the anchor orientation, and none of these failures is visible from the prompt.
We therefore treat rewrites as proposals rather than substitutions, accepting one only when the generated image shows a concrete improvement.
This yields a training-free strategy that requires no modification to the generation model.

\noindent \textbf{Candidate generation.}
Given a FoR prompt $p$, we generate a reference image $I_0$ from the original prompt, and a language model produces ${K}$ rewrites $\{p_1,\ldots,p_K\}$ that preserve the objects, camera view, anchor orientation, and intended relation while stating the required frame conversion more explicitly.
Each rewrite yields one candidate image $I_k$ under the same T2I model and inference settings as $I_0$.

% [ZZY] 把 checklist 包括整个training-free方法用到的prompt 放附录，这里cite一下？ okk，不过checklist和appendix单独交的，我把要放附录的都加一下
\noindent\textbf{Visual verification.}
From the original prompt we derive a checklist covering the required objects, the anchor orientation, and the target placement, and a vision-language model (Qwen3.6 27B) compares each candidate $I_k$ against $I_0$ on this checklist.
A candidate is accepted only if it improves the target relation without dropping objects or violating the orientation constraint; if none is accepted, we keep $I_0$.
Checklist construction and all prompts used are given in the supplementary material.

\noindent\textbf{Reference-anchored selection.}
We compare each candidate against $I_0$ rather than ranking candidates directly, because unconstrained ranking tends to reward visual quality over spatial correctness.
The gate asks only whether a rewrite yields a concrete improvement in the required relation, which makes the method conservative: it replaces $I_0$ only when the evidence favors doing so.

\section{Experiments}

\subsection{Experimental Setup}

We cover both closed-source models and open-source models.
All models are evaluated on the same set of Cam prompts and FoR prompts using the metrics defined above.
Generated images are scored with an automatic evaluator combining OWL-ViT, Segment Anything Model 3.1 (SAM3.1) \cite{carion2025sam3segmentconcepts}, Depth Anything 3 (DA3) \cite{depthanything3}, and a VLM judge. 
The evaluator checks object presence and count, localizes the requested objects, determines their image-frame relations, and assesses anchor orientation. Its reliability is evaluated against human annotations.

\paragraph{Metrics.}
We report geometry accuracy, orientation accuracy, and final accuracy.
Geometry accuracy measures whether all required target relations are satisfied in the image frame, independently of whether the requested anchor orientations are rendered correctly. Orientation accuracy measures whether all required anchor orientations are satisfied.
Final accuracy requires both geometry and orientation to be correct. For prompts with multiple constraints, all constraints must be satisfied. Avg. denotes the macro average over L1--L3.
Because the Cam and FoR prompts in each pair specify the same ground-truth layout, we define
$Gap =\mathrm{Acc}_{\mathrm{Cam}}-\mathrm{Acc}_{\mathrm{FoR}}$.
This difference estimates the additional difficulty introduced when the same layout is expressed through an anchor-centered relation.
Missing required objects and additional instances of a category requested exactly once are scored as incorrect.

\subsection{Main Results}

Table~\ref{tab:main_results} reports spatial accuracy by benchmark task level. 
\begin{table}[t]
\centering
\tiny
\setlength{\tabcolsep}{2pt}
\resizebox{\columnwidth}{!}{
\begin{tabular}{lrrrrrrrr}
\toprule
& \multicolumn{2}{c}{Avg.} & \multicolumn{2}{c}{L1} & \multicolumn{2}{c}{L2} & \multicolumn{2}{c}{L3} \\
\cmidrule(lr){2-3}\cmidrule(lr){4-5}\cmidrule(lr){6-7}\cmidrule(lr){8-9}
Model & Cam & FoR & Cam & FoR & Cam & FoR & Cam & FoR \\
\midrule
\multicolumn{9}{l}{\textit{Closed-source models}} \\
GPT-Image-2 & 68.2 & 38.2 & 75.1 & 50.9 & 73.4 & 51.0 & 56.1 & 12.8 \\
Seedream 5.0 & 64.1 & 44.3 & 73.1 & 55.1 & 68.0 & 54.1 & 51.2 & 23.6 \\
Gemini 3 Pro Image & 53.5 & 35.9 & 65.2 & 45.8 & 58.5 & 47.5 & 36.9 & 14.4 \\
Gemini 3.1 Flash Image & 52.0 & 29.5 & 62.8 & 38.6 & 58.1 & 38.8 & 35.1 & 11.2 \\
Qwen-Image 2.0 & 48.8 & 25.4 & 59.0 & 35.4 & 58.0 & 32.8 & 29.3 & 8.1 \\
Seedream 4.0 & 44.3 & 23.1 & 60.3 & 33.2 & 55.8 & 32.2 & 16.8 & 3.8 \\
Gemini 2.5 Flash Image & 40.9 & 19.6 & 50.6 & 27.4 & 47.7 & 27.3 & 24.4 & 4.1 \\
GPT-Image-1 & 33.3 & 17.0 & 43.1 & 23.3 & 37.3 & 21.3 & 19.6 & 6.3 \\
\midrule
\multicolumn{9}{l}{\textit{Open-source models}} \\
Qwen-Image-2512 & 33.9 & 17.9 & 52.8 & 30.3 & 37.5 & 21.4 & 11.3 & 2.1 \\
HunyuanImage 3.0 & 24.7 & 13.2 & 39.9 & 22.5 & 27.0 & 14.6 & 7.3 & 2.6 \\
Z-Image Turbo & 26.1 & 13.1 & 38.6 & 21.2 & 31.6 & 15.5 & 8.0 & 2.4 \\
ERNIE-Image & 17.5 & 9.0 & 33.7 & 15.9 & 13.1 & 9.6 & 5.8 & 1.5 \\
Qwen-Image & 17.4 & 9.3 & 31.5 & 17.6 & 15.8 & 9.6 & 4.9 & 0.6 \\
UniPic v2 & 10.8 & 7.8 & 19.1 & 14.8 & 10.9 & 7.5 & 2.4 & 1.1 \\
Infinity & 9.6 & 7.8 & 19.3 & 15.0 & 7.9 & 7.7 & 1.7 & 0.6 \\
OmniGen2 & 8.2 & 5.2 & 14.6 & 9.2 & 8.6 & 5.6 & 1.5 & 0.9 \\
Janus-Pro-7B & 7.1 & 5.5 & 14.2 & 11.4 & 6.4 & 4.3 & 0.9 & 0.9 \\
FLUX.1-dev & 6.9 & 5.4 & 15.2 & 11.4 & 4.7 & 3.9 & 0.9 & 1.1 \\
BAGEL-7B-MoT & 6.1 & 4.8 & 11.8 & 8.4 & 5.8 & 4.7 & 0.6 & 1.3 \\
Show-o & 5.4 & 4.5 & 12.2 & 9.7 & 3.0 & 3.2 & 0.9 & 0.6 \\
SD3.5 Medium & 3.7 & 2.5 & 8.4 & 6.0 & 2.4 & 1.3 & 0.4 & 0.2 \\
FLUX.1-Fill-dev & 0.2 & 0.1 & 0.6 & 0.0 & 0.0 & 0.4 & 0.0 & 0.0 \\
\bottomrule
\end{tabular}
}
\caption{Accuracy by task level and prompt type. Cam denotes camera view prompts, while FoR denotes prompts that describe the same target layout through an oriented anchor object. Avg. reports the macro average over L1--L3.}
\label{tab:main_results}
\end{table}
The closed-source group includes Seedream 5.0 Pro~\citep{bytedance2026seedream5pro}, GPT-Image-2~\citep{openai2026gptimage2}, Gemini 3 Pro Image~\citep{google2025gemini3proimage}, Gemini 3.1 Flash Image~\citep{google2026gemini31flashimage}, Qwen-Image 2.0~\citep{zhao2026qwenimage2}, Seedream 4.0~\citep{chen2025seedream4}, Gemini 2.5 Flash Image~\citep{google2025gemini25flashimage}, and GPT-Image-1~\citep{openai2025gptimage1}.
The open-source group includes Qwen-Image-2512~\citep{wu2025qwenimagetechnicalreport}, HunyuanImage 3.0~\citep{cao2025hunyuanimage3}, Z-Image Turbo~\citep{zimage2025}, Qwen-Image~\citep{wu2025qwenimage}, ERNIE-Image~\citep{baidu2026ernieimage}, UniPic v2~\citep{wei2025unipic2}, Infinity~\citep{han2024infinity}, OmniGen2~\citep{wu2025omnigen2}, Janus-Pro-7B~\citep{chen2025januspro}, FLUX.1-dev~\citep{flux2024}, BAGEL-7B-MoT~\citep{deng2025bagel}, Show-o~\citep{xie2024showo}, SD3.5 Medium~\citep{esser2024sd3}, and FLUX.1-Fill-dev~\citep{flux2024}.

\begin{table*}[t]
\centering
\tiny
\setlength{\tabcolsep}{1.7pt}
\resizebox{\textwidth}{!}{
\begin{tabular}{l*{19}{r}}
\toprule
& \multicolumn{8}{c}{Geometry}
& \multicolumn{8}{c}{Orientation}
& \multicolumn{3}{c}{\multirow{2}{*}{Avg.}} \\
\cmidrule(lr){2-9}\cmidrule(lr){10-17}
& \multicolumn{2}{c}{L1}
& \multicolumn{2}{c}{L2}
& \multicolumn{2}{c}{L3}
& \multicolumn{2}{c}{Avg.}
& \multicolumn{2}{c}{L1}
& \multicolumn{2}{c}{L2}
& \multicolumn{2}{c}{L3}
& \multicolumn{2}{c}{Avg.}
& \multicolumn{3}{c}{} \\
\cmidrule(lr){2-3}\cmidrule(lr){4-5}
\cmidrule(lr){6-7}\cmidrule(lr){8-9}
\cmidrule(lr){10-11}\cmidrule(lr){12-13}
\cmidrule(lr){14-15}\cmidrule(lr){16-17}
\cmidrule(lr){18-20}
Model
& Cam & FoR & Cam & FoR & Cam & FoR & Cam & FoR
& Cam & FoR & Cam & FoR & Cam & FoR & Cam & FoR
& Cam & FoR & Gap \\
\midrule

\multicolumn{20}{l}{\textit{Closed-source models}} \\
Seedream 5.0
& 94.0 & 68.1 & 85.7 & 64.7 & 84.5 & 42.4 & 88.1 & 58.4
& 77.5 & 79.3 & 79.7 & 81.2 & 59.4 & 44.1 & 72.2 & 68.2
& 64.1 & 44.3 & 19.8 \\
GPT-Image-2
& 87.7 & 60.4 & 84.8 & 59.0 & 80.4 & 27.2 & 84.3 & 48.9
& 79.5 & 80.6 & 80.1 & 83.7 & 66.8 & 33.8 & 75.4 & 66.0
& 68.2 & 38.2 & 30.0 \\
Gemini 3 Pro Image
& 90.7 & 65.9 & 83.7 & 65.4 & 77.7 & 38.9 & 84.0 & 56.8
& 71.1 & 67.6 & 67.3 & 68.9 & 43.3 & 32.0 & 60.6 & 56.2
& 53.5 & 35.9 & 17.6 \\
Gemini 3.1 Flash Image
& 89.2 & 51.8 & 78.6 & 54.5 & 75.5 & 27.2 & 81.1 & 44.5
& 69.6 & 71.8 & 72.2 & 69.5 & 44.4 & 27.6 & 62.1 & 56.3
& 52.0 & 29.5 & 22.5 \\
Qwen-Image 2.0
& 90.6 & 53.4 & 87.8 & 49.3 & 83.3 & 26.6 & 87.2 & 43.1
& 64.2 & 65.2 & 64.2 & 64.9 & 34.7 & 23.1 & 54.4 & 51.1
& 48.8 & 25.4 & 23.4 \\
Seedream 4.0
& 94.0 & 47.3 & 88.5 & 47.7 & 77.7 & 21.6 & 86.7 & 38.8
& 64.8 & 69.8 & 62.2 & 67.1 & 19.4 & 10.2 & 48.8 & 49.0
& 44.3 & 23.1 & 21.2 \\
Gemini 2.5 Flash Image
& 85.0 & 44.7 & 79.8 & 43.9 & 75.2 & 24.6 & 80.0 & 37.7
& 55.1 & 51.3 & 58.9 & 52.5 & 32.2 & 18.4 & 48.7 & 40.7
& 40.9 & 19.6 & 21.3 \\
GPT-Image-1
& 88.4 & 47.5 & 75.7 & 42.5 & 71.6 & 28.4 & 78.6 & 39.4
& 49.0 & 53.0 & 48.0 & 48.0 & 27.6 & 16.4 & 41.5 & 39.1
& 33.3 & 17.0 & 16.4 \\

\midrule
\multicolumn{20}{l}{\textit{Open-source models}} \\
Qwen-Image-2512
& 81.8 & 44.4 & 63.4 & 34.5 & 57.8 & 17.8 & 67.7 & 32.2
& 62.4 & 66.7 & 56.1 & 62.7 & 18.8 & 10.9 & 45.8 & 46.8
& 33.9 & 17.9 & 15.9 \\
HunyuanImage 3.0
& 79.4 & 46.4 & 56.3 & 33.4 & 47.8 & 21.0 & 61.2 & 33.6
& 50.9 & 52.6 & 48.0 & 49.0 & 14.3 & 11.3 & 37.7 & 37.7
& 24.7 & 13.2 & 11.5 \\
Z-Image Turbo
& 85.7 & 45.4 & 76.1 & 41.7 & 54.3 & 21.8 & 72.0 & 36.3
& 44.3 & 46.0 & 41.5 & 39.5 & 13.2 & 9.0 & 33.0 & 31.5
& 26.1 & 13.1 & 13.1 \\
Qwen-Image
& 73.0 & 37.8 & 40.9 & 24.6 & 36.4 & 13.9 & 50.1 & 25.4
& 42.3 & 47.6 & 37.0 & 39.2 & 9.0 & 6.0 & 29.4 & 30.9
& 17.4 & 9.3 & 8.1 \\
ERNIE-Image
& 63.7 & 38.4 & 27.8 & 19.7 & 29.6 & 12.0 & 40.4 & 23.4
& 48.7 & 45.1 & 39.0 & 39.8 & 12.4 & 7.9 & 33.4 & 30.9
& 17.5 & 9.0 & 8.5 \\
UniPic v2
& 71.5 & 41.6 & 37.7 & 24.8 & 26.8 & 17.3 & 45.3 & 27.9
& 26.8 & 29.6 & 26.6 & 27.2 & 3.0 & 3.2 & 18.8 & 20.0
& 10.8 & 7.8 & 3.0 \\
Infinity
& 62.7 & 40.8 & 28.7 & 24.0 & 27.2 & 15.4 & 39.5 & 26.7
& 29.6 & 31.1 & 28.1 & 33.2 & 3.9 & 3.0 & 20.5 & 22.4
& 9.6 & 7.8 & 1.9 \\
OmniGen2
& 49.4 & 29.6 & 26.6 & 15.6 & 19.3 & 9.6 & 31.7 & 18.3
& 26.0 & 27.7 & 19.7 & 22.7 & 4.7 & 4.1 & 16.8 & 18.1
& 8.2 & 5.2 & 3.0 \\
Janus-Pro-7B
& 42.7 & 38.0 & 15.2 & 14.6 & 13.9 & 12.2 & 23.9 & 21.6
& 28.5 & 29.6 & 26.8 & 29.1 & 5.4 & 5.8 & 20.2 & 21.5
& 7.1 & 5.5 & 1.6 \\
FLUX.1-dev
& 50.0 & 32.8 & 18.2 & 13.5 & 18.0 & 9.9 & 28.7 & 18.7
& 29.2 & 30.0 & 24.4 & 29.8 & 3.2 & 4.3 & 18.9 & 21.4
& 6.9 & 5.4 & 1.5 \\
BAGEL-7B-MoT
& 42.9 & 33.3 & 18.8 & 13.5 & 12.8 & 11.1 & 24.9 & 19.3
& 23.6 & 25.5 & 24.6 & 25.7 & 3.4 & 5.1 & 17.2 & 18.8
& 6.1 & 4.8 & 1.3 \\
Show-o
& 41.6 & 35.2 & 13.9 & 14.1 & 14.6 & 9.2 & 23.4 & 19.5
& 27.7 & 25.3 & 19.1 & 22.3 & 4.5 & 4.1 & 17.1 & 17.2
& 5.4 & 4.5 & 0.9 \\
SD3.5 Medium
& 41.0 & 34.1 & 10.9 & 7.3 & 10.5 & 9.2 & 20.8 & 16.9
& 17.2 & 17.0 & 18.2 & 14.8 & 2.8 & 1.3 & 12.7 & 11.0
& 3.7 & 2.5 & 1.2 \\
FLUX.1-Fill-dev
& 24.7 & 25.8 & 4.7 & 5.8 & 6.0 & 6.0 & 11.8 & 12.5
& 1.9 & 0.9 & 0.6 & 0.6 & 0.0 & 0.0 & 0.9 & 0.5
& 0.2 & 0.1 & 0.1 \\
\bottomrule
\end{tabular}
}
\caption{
Component-wise model breakdown by task level.
Cam denotes camera-frame prompts.
Geometry measures whether the target spatial relation is satisfied, while
Orientation measures whether the anchor object's facing direction is correct.
The final Avg. columns report strict accuracy requiring both components to be
correct.
Gap is Cam minus FoR in percentage points and is computed before rounding.
All Avg. columns are macro-averaged over L1--L3.
}
\label{tab:component_breakdown_by_model}
\vspace{-0.2cm}
\end{table*}

% 一些有的没的的分析

Table~\ref{tab:main_results} reports final accuracy for each model and task level. Every model obtains lower average accuracy on FoR prompts than on its paired Cam prompts. Averaged equally across the 22 models, final accuracy decreases from 26.5\% on Cam to 15.4\% on FoR, an 11.1-point difference. The best FoR result is only 44.3\%, showing that substantial room for improvement remains even among the strongest models.

The gap persists in models with strong image-frame spatial control.
Among closed-source models, average accuracy decreases from 50.6\% to 29.1\%; among open-source models, it decreases from 12.7\% to 7.6\%. The smaller absolute gap in the latter group should be interpreted in light of its much lower Cam baseline and the resulting floor effect.
GPT-Image-2 achieves the highest Cam accuracy at 68.2\% but falls to 38.2\% on FoR prompts. Seedream 5.0 obtains the highest FoR accuracy at 44.3\%, while still showing a 19.8-point gap. These results indicate that strong performance on image-frame descriptions does not reliably transfer to anchor-centered descriptions of the same layout.

% Both prompt types become more difficult at higher task levels. Mean Cam/FoR accuracy is 36.4\%/22.9\% at L1 and 28.7\%/18.8\% at L2, before falling to 14.4\%/4.6\% at L3.
% At L3, mean FoR accuracy is less than one third of the corresponding Cam accuracy, compared with roughly two thirds at L1 and L2.
% Although the absolute gap is not monotonic because many weaker models approach the performance floor, the pronounced relative decline at L3 shows that generation is particularly fragile when target placement depends on an additional orientation constraint.
Both prompt types degrade at higher task levels, but not at the same rate.
Mean Cam/FoR accuracy is 35.3\%/21.5\% at L1 and 28.7\%/19.6\% at L2, before falling to 14.4\%/4.5\% at L3.
FoR accuracy thus retains roughly two thirds of the corresponding Cam accuracy at L1 and L2, but less than one third at L3.
The absolute gap is not monotonic, since many weaker models approach the floor at L3, but the relative collapse is consistent across the strongest models.
This decline is steeper than repeated single-step resolution would predict. If the two frame resolutions required at L3 failed independently, L3 accuracy would be approximately the square of L1 accuracy: for the closed-source models this predicts 15.0\% but the observed value is 10.5\%, whereas the same estimate on Cam prompts is far closer (37.5\% predicted, 33.7\% observed).
Chained frame resolution therefore costs more than two independent conversions, and the excess is specific to the FoR condition.

\subsection{Component-Level Results}

Table~\ref{tab:component_breakdown_by_model} separates target placement from anchor orientation. Because paired Cam and FoR prompts share the same orientation requirements, a degradation specifically associated with the FoR description should appear primarily in geometry accuracy. This is what we observe: averaged across models, geometry accuracy falls
from 55.1\% on Cam prompts to 31.8\% on FoR, a difference of 23.3 points. Orientation accuracy, changes only from 35.7\% to 34.3\%. The additional FoR difficulty is therefore expressed mainly through incorrect image-frame placement of the target objects.

The small difference in orientation accuracy should not be interpreted as strong orientation generation. Its absolute value remains low under both prompt types and falls at L3 from 19.4\% on Cam prompts to 12.8\% on FoR prompts. Anchor orientation is therefore a general difficulty shared by both conditions, whereas target placement accounts for most of the Cam--FoR gap. Since final accuracy requires both components to be correct, their combined failure is especially severe at L3.

\subsection{Training-free Mitigation Results}
\begin{table}[t]
\centering
\small
\setlength{\tabcolsep}{5pt}
\begin{tabular}{lcccc}
\toprule
Method & L1 & L2 & L3 & Avg. \\
\midrule
No rewriting & 30.0 & 31.7 & 13.3 & 25.0 \\
Naive rewriting & 37.5 & 35.8 & 10.8 & 28.0 \\
DSG \cite{dsg} &  29.2 & 35.0 & 6.7 & 23.6 \\
TIFA \cite{tifa} & 33.3 & 35.8 & 5.8 & 25.0 \\
T2I Copilot \cite{T2Icopilot} & 33.3 & 39.2 & 10.8 & 27.8 \\
VisualPrompter \cite{visualprompter} & 33.3 & 38.3 & 8.3 & 26.7 \\
VisionDirector \cite{visiondirector} & 34.2 & 36.7 & 10.8 & 27.2 \\
Ours & 34.2 & 38.3 & 15.0 & 29.2 \\
\bottomrule
\end{tabular}
\caption{
Comparison of training-free enhancement strategies on FoR prompts. All methods are evaluated with three backbones. L1, L2, and L3 denote the three FoR difficulty levels, and Avg. reports the macro average across levels.
}
\label{tab:training_free_enhancement}
\end{table}

We compare our method with six training-free baselines. No rewriting directly uses the original FoR prompt.
Naive rewrite generates rewritten prompts without visual verification. We use Qwen3.6-27B for prompt rewrite.
DSG~\citep{dsg} and TIFA~\citep{tifa} use question-based decomposition, while T2I Copilot~\citep{T2Icopilot}, VisualPrompter~\citep{visualprompter}, and VisionDirector~\citep{visiondirector} use prompt planning or visual feedback.
We adapt all methods to the same generation budget.

We evaluate each method on a 10\% subset of \bench, containing 120 FoR prompts.
Experiments use three T2I backbones: Z-Image Turbo, Seedream 4.0, and Seedream 5.0.
Our method achieves the highest average accuracy, improving the baseline from 25.0\% to 29.2\%.
It also provides the best L3 result, increasing accuracy from 13.3\% to 15.0\%.
Every method improve L1 or L2 but reduce L3 accuracy.
This result suggests that visual verification is particularly important when prompt rewriting must preserve chained orientation and placement constraints.

% \subsection{Evaluator Reliability}
% \label{sec:evaluator_validation}

% We validate the automatic evaluator against human annotations on 120 prompt pairs, covering 10.0\% of the benchmark.
% Four annotators assess object presence, object count, target placement, anchor orientation, and final correctness.
% We separately evaluate the object detection stage and the complete evaluation pipeline.

% Table~\ref{tab:evaluator_validation} validates object grounding and final evaluation separately.
% In panel (a), OWL-ViT with SAM3.1 grounds 98.9\% of required objects and completely grounds 95.0\% of prompt pairs.
% In panel (b), our full evaluator combines OWL-ViT and SAM3.1 for grounding, DA3 for depth estimation, and Qwen3.6-27B for orientation judgment, achieving 87.1\% accuracy and an F1 score of 88.5 against human annotations.
% Unresolved samples are conservatively scored as incorrect.
\subsection{Evaluator Reliability}
\label{sec:evaluator_validation}

We validate the automatic evaluator against human annotations on 120 prompt pairs, covering 10.0\% of the benchmark.
Four annotators assess object presence, object count, target placement, anchor orientation, and final correctness.
We separately evaluate object detection and the complete evaluation pipeline.
This separation distinguishes grounding failures from errors in spatial and orientation judgments.

Table~\ref{tab:evaluator_validation} validates object grounding and final evaluation separately.
In panel (a), OWL-ViT with SAM3.1 grounds 98.9\% of required objects and completely grounds 95.0\% of prompt pairs.
In panel (b), our full evaluator combines OWL-ViT and SAM3.1 for grounding, DA3 for depth estimation, and Qwen3.6-27B for orientation judgment, achieving 87.1\% accuracy and an F1 score of 88.5 against human annotations.
This exceeds the 65.0\% accuracy of the standalone VLM judge.
Unresolved samples are conservatively scored as incorrect.

\begin{table}[t]
\centering
\small
\setlength{\tabcolsep}{3pt}
\renewcommand{\arraystretch}{1.08}

\begin{tabular*}{\columnwidth}{
@{\extracolsep{\fill}}lccc@{}}
\toprule
\multicolumn{4}{c}{\textbf{(a) Object grounding coverage}} \\
\cmidrule(lr){1-4}
Grounding module
& Obj. rec.
& Count acc.
& Pair comp. \\
\midrule
OWL-ViT
& 95.8
& 89.6
& 84.2 \\
OWL-ViT + SAM3.1
& \textbf{98.9}
& \textbf{97.1}
& \textbf{95.0} \\
\bottomrule
\end{tabular*}

\vspace{5pt}

\begin{tabular*}{\columnwidth}{
@{\extracolsep{\fill}}lcccc@{}}
\toprule
\multicolumn{5}{c}{\textbf{(b) Agreement with human judgments}} \\
\cmidrule(lr){1-5}
Evaluator
& Acc.
& Prec.
& Rec.
& F1 \\
\midrule
VLM judge
& 65.0
& 66.1
& 79.6
& 72.2 \\
OWL-ViT pipeline
& 79.6
& 90.0
& 72.3
& 80.2 \\
OWL-ViT + SAM3.1 pipeline
& \textbf{87.1}
& \textbf{90.2}
& \textbf{86.9}
& \textbf{88.5} \\
\bottomrule
\end{tabular*}

\caption{
Evaluator validation on 120 prompt pairs.
(a) Object grounding coverage; pair completeness requires all objects to be grounded in both Cam and FoR images.
(b) Agreement with human final-correctness annotations.
All values are percentages; Qwen3.6-27B judges orientation.
}
\label{tab:evaluator_validation}
\end{table}

\begin{table}[t]
\centering
\small
\setlength{\tabcolsep}{5pt}
\begin{tabular*}{\columnwidth}{@{\extracolsep{\fill}}lrrr@{}}
\toprule
Condition & Cam & FoR & Gap \\
\midrule
\multicolumn{4}{l}{\textit{Relation type (relation-level accuracy)}} \\
Front & 63.2 & 41.0 & 22.2 \\
Back  & 64.5 & 46.5 & 18.0 \\
Left  & 63.2 & 37.2 & 26.0 \\
Right & 63.0 & 36.4 & 26.6 \\
\midrule
\multicolumn{4}{l}{\textit{Camera view (geometry accuracy)}} \\
Eye-level & 53.8 & 32.3 & 21.5 \\
Top-down  & 56.5 & 31.4 & 25.0 \\
\midrule
\multicolumn{4}{l}{\textit{Lateral frame mapping (relation-level accuracy)}} \\
Preserved & 66.0 & 60.6 & 5.3 \\
Axis/depth remapping & 59.6 & 40.1 & 19.5 \\
Reversed & 68.5 & 18.8 & 49.7 \\
\bottomrule
\end{tabular*}
\caption{
Accuracy by relation, camera view, and frame mapping.
Preserved, remapped, and reversed indicate whether anchor-relative left/right maps to the same side, another axis or depth, or the opposite side.
Results are macro-averaged over models; Gap is Cam minus FoR in percentage points.
}
\label{tab:analysis_breakdown}
\vspace{-0.4cm}
\end{table}

\section{Analysis}

To identify the sources of the Cam--FoR gap, we analyze performance by relation type, camera view, and frame mapping.
Unless noted otherwise, results are macro-averaged across models.
Relation-level accuracy evaluates only the anchor--target relation rather than all spatial constraints in a prompt.

\begin{figure}[t]
    \centering
    \includegraphics[width=1\linewidth]{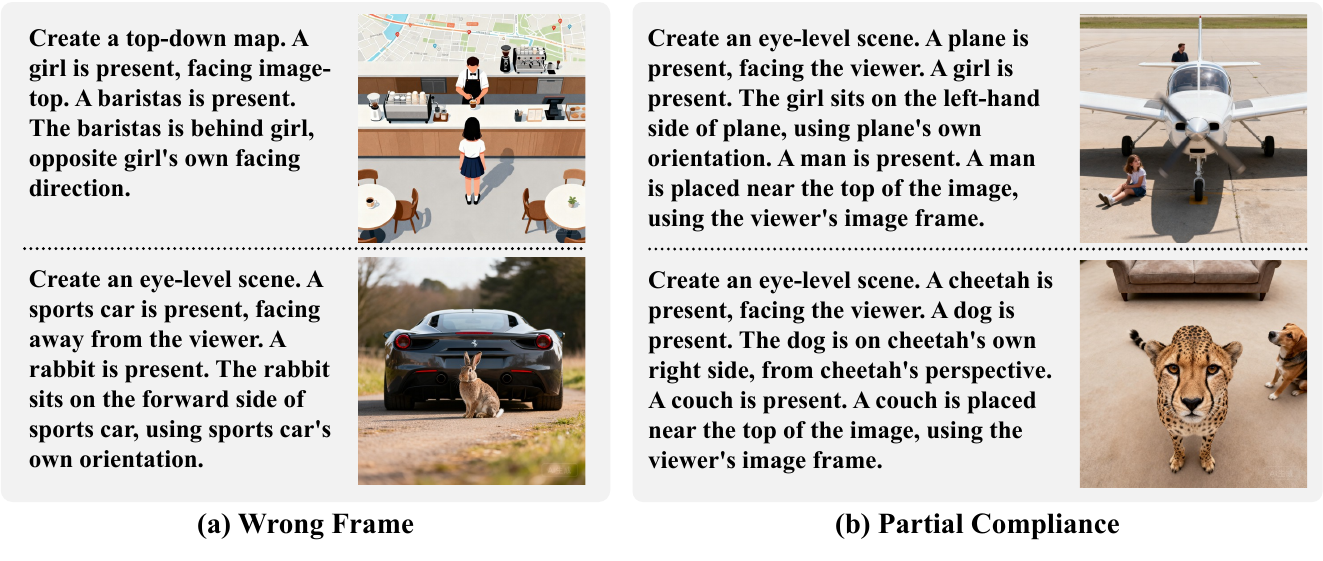}
    \caption{Representative failure patterns under FoR prompts.
(a) \textbf{Wrong Frame Interpretation}: the model resolves an object-relative direction in viewer-centered coordinates, placing the barista and rabbit on the wrong side of their oriented anchors.
(b) \textbf{Partial Compliance}: the model satisfies the viewer-centered constraints for the man and couch, but violates the object-relative constraints for the girl and dog.}
    \vspace{-0.3cm}
    \label{fig:failure_patterns}
\end{figure}

\subsection{Performance by Relation Type}

Table~\ref{tab:analysis_breakdown} first separates the primary spatial relation into front, back, left, and right. Cam relation-level accuracy is nearly constant across the four categories, ranging only from 63.0\% to 64.5\%. The corresponding FoR accuracy is lower for every relation and varies more substantially, from 36.4\% to 46.5\%. Left and right produce the largest Cam--FoR gaps, at 26.0 and 26.6 percentage points, respectively, compared with 22.2 points for front and 18.0 points for back. The stable Cam results indicate that this contrast is not primarily caused by the image-frame target positions associated with different relation types. Instead, the larger lateral gaps suggest that models have particular difficulty applying an anchor's oriented left--right axis during generation.

This pattern is even clearer among closed-source models, for which the gaps for left and right are 45.4 and 41.5 points, compared with 30.3 and 24.3 points for front and back. The smaller gaps among open-source models should be interpreted in light of their lower Cam baselines, which leave less room for an additional absolute decrease.

\subsection{Viewpoint and Frame Alignment}

The FoR deficit appears in both camera views. Geometry accuracy under FoR prompts is similar for eye-level and top-down scenes (32.3\% and 31.4\%), while the corresponding Cam accuracies are 53.8\% and 56.5\%. Thus, neither view alone accounts for the degradation.

Anchor orientation is more informative when expressed through the frame mapping it induces. We therefore group anchor-relative left/right relations by whether the required side is preserved in the image frame, mapped to a different image or depth axis, or reversed. When the two frames agree, FoR relation-level accuracy approaches the Cam baseline (60.6\% versus 66.0\%). The gap increases to 19.5 points under axis or depth remapping and reaches 49.7 points under reversal, where FoR accuracy falls to 18.8\% despite a Cam accuracy of 68.5\%. Because the Cam condition remains strong on these layouts, the reversal failure cannot be explained by output-layout difficulty alone; it instead identifies the reference-frame conversion as the main source of error. All 22 models show a positive Cam–FoR gap under reversal. Raw differences between anchor orientations largely reflect these mappings and are less diagnostic in isolation.

\subsection{Failure Patterns}

\noindent\textbf{Wrong frame interpretation.}
Models often apply an object-relative direction directly in viewer-centered image coordinates, as shown in Figure~\ref{fig:failure_patterns}(a).
Among reversed left/right cases in which all required objects are localizable, 74.1\% of FoR generations place the target on the literal image side named by the object-relative relation, compared with 16.7\% under the paired Cam prompts.
This behavior indicates that models frequently ignore the anchor's rendered orientation.

\noindent\textbf{Partial compliance.}
In prompts containing multiple spatial constraints, models often satisfy at least one relation but not the complete layout.
Across L2 and L3, partial compliance occurs in approximately 32\% of FoR generations, compared with 18\% under Cam prompts.
The examples in Figure~\ref{fig:failure_patterns}(b) satisfy the viewer-centered constraint while violating the object-relative relation.

\section{Conclusion}

We introduce \bench, a controlled benchmark that tests whether T2I models can generate layouts from frame-dependent spatial descriptions, using matched Cam and FoR prompts that specify the same target layout.
Across 22 models, the loss is concentrated in a specific condition: when the anchor's left--right axis is reversed relative to the image, FoR accuracy drops to 18.8\% while the paired Cam prompts reach 68.5\% on the same layouts.
Our VLM-gated rewriting provides a modest improvement, showing that visual feedback can partially mitigate these errors without model updates.
The remaining gap motivates future work on orientation-aware representations and training objectives that explicitly connect linguistic reference frames to image coordinates.

% TODO: Summarize the benchmark, the diagnostic gaps, and the main empirical finding without overclaiming beyond the evaluated prompt families and image-coordinate layout setting.

\bibliography{paper}

@book{levinson2003space,
  author    = {Levinson, Stephen C.},
  title     = {Space in Language and Cognition: Explorations in Cognitive Diversity},
  year      = {2003},
  publisher = {Cambridge University Press},
  address   = {Cambridge, UK}
}

@misc{gokhale2022benchmarking,
  title         = {Benchmarking Spatial Relationships in Text-to-Image Generation},
  author        = {Gokhale, Tejas and Palangi, Hamid and Nushi, Besmira and Vineet, Vibhav and Horvitz, Eric and Kamar, Ece and Baral, Chitta and Yang, Yezhou},
  year          = {2022},
  eprint        = {2212.10015},
  archivePrefix = {arXiv},
  primaryClass  = {cs.CV}
}

@misc{huang2023t2icompbench,
  title         = {T2I-CompBench++: An Enhanced and Comprehensive Benchmark for Compositional Text-to-image Generation},
  author        = {Huang, Kaiyi and Duan, Chengqi and Sun, Kaiyue and Xie, Enze and Li, Zhenguo and Liu, Xihui},
  year          = {2025},
  eprint        = {2307.06350},
  archivePrefix = {arXiv},
  primaryClass  = {cs.CV}
}

@misc{ghosh2023geneval,
  title         = {GenEval: An Object-Focused Framework for Evaluating Text-to-Image Alignment},
  author        = {Ghosh, Dhruba and Hajishirzi, Hanna and Schmidt, Ludwig},
  year          = {2023},
  eprint        = {2310.11513},
  archivePrefix = {arXiv},
  primaryClass  = {cs.CV}
}

@misc{premsri2025forest,
  title         = {FoREST: Frame of Reference Evaluation in Spatial Reasoning Tasks},
  author        = {Premsri, Tanawan and Kordjamshidi, Parisa},
  year          = {2025},
  eprint        = {2502.17775},
  archivePrefix = {arXiv},
  primaryClass  = {cs.CL}
}

@misc{li2025viewspatial,
  title         = {ViewSpatial-Bench: Evaluating Multi-perspective Spatial Localization in Vision-Language Models},
  author        = {Li, Dingming and Li, Hongxing and Wang, Zixuan and Yan, Yuchen and Zhang, Hang and Chen, Siqi and Hou, Guiyang and Jiang, Shengpei and Zhang, Wenqi and Shen, Yongliang and Lu, Weiming and Zhuang, Yueting},
  year          = {2025},
  eprint        = {2505.21500},
  archivePrefix = {arXiv},
  primaryClass  = {cs.CV}
}

@misc{wang2025genspace,
  title         = {GenSpace: Benchmarking Spatially-Aware Image Generation},
  author        = {Wang, Zehan and Xu, Jiayang and Zhang, Ziang and Pang, Tianyu and Du, Chao and Zhao, Hengshuang and Zhao, Zhou},
  year          = {2025},
  eprint        = {2505.24870},
  archivePrefix = {arXiv},
  primaryClass  = {cs.CV}
}

@misc{wang2026spatialgeneval,
  title         = {Everything in Its Place: Benchmarking Spatial Intelligence of Text-to-Image Models},
  author        = {Wang, Zengbin and Hu, Xuecai and Wang, Yong and Xiong, Feng and Zhang, Man and Chu, Xiangxiang},
  year          = {2026},
  eprint        = {2601.20354},
  archivePrefix = {arXiv},
  primaryClass  = {cs.CV}
}

@misc{wang2026allocentric,
  title         = {Allocentric Perceiver: Disentangling Allocentric Reasoning from Egocentric Visual Priors via Frame Instantiation},
  author        = {Wang, Hengyi and Zhang, Ruiqiang and Liu, Chang and Wang, Guanjie and Ma, Zehua and Fang, Han and Zhang, Weiming},
  year          = {2026},
  eprint        = {2602.05789},
  archivePrefix = {arXiv},
  primaryClass  = {cs.CV}
}

@misc{zhao2026qwenimage2,
  title         = {Qwen-Image-2.0 Technical Report},
  author        = {{Qwen Team}},
  year          = {2026},
  eprint        = {2605.10730},
  archivePrefix = {arXiv},
  primaryClass  = {cs.CV}
}

@misc{cao2025hunyuanimage3,
  title         = {HunyuanImage 3.0 Technical Report},
  author        = {{HunyuanImage Team}},
  year          = {2025},
  eprint        = {2509.23951},
  archivePrefix = {arXiv},
  primaryClass  = {cs.CV}
}

@misc{zimage2025,
  title         = {Z-Image: An Efficient Image Generation Foundation Model with Single-Stream Diffusion Transformer},
  author        = {{Z-Image Team}},
  year          = {2025},
  eprint        = {2511.22699},
  archivePrefix = {arXiv},
  primaryClass  = {cs.CV}
}

@misc{wu2025omnigen2,
  title         = {OmniGen2: Exploration to Advanced Multimodal Generation},
  author        = {Wu, Chenyuan and Zheng, Pengfei and Yan, Ruiran and Xiao, Shitao and Luo, Xin and Wang, Yueze and Li, Wanli and Jiang, Xiyan and Liu, Yexin and Zhou, Junjie and Liu, Ze and Xia, Ziyi and Li, Chaofan and Deng, Haoge and Wang, Jiahao and Luo, Kun and Zhang, Bo and Lian, Defu and Wang, Xinlong and Wang, Zhongyuan and Huang, Tiejun and Liu, Zheng},
  year          = {2025},
  eprint        = {2506.18871},
  archivePrefix = {arXiv},
  primaryClass  = {cs.CV}
}

@misc{deng2025bagel,
  title         = {Emerging Properties in Unified Multimodal Pretraining},
  author        = {Deng, Chaorui and Zhu, Deyao and Li, Kunchang and Gou, Chenhui and Li, Feng and Wang, Zeyu and Zhong, Shu and Yu, Weihao and Nie, Xiaonan and Song, Ziang and Shi, Guang and Fan, Haoqi},
  year          = {2025},
  eprint        = {2505.14683},
  archivePrefix = {arXiv},
  primaryClass  = {cs.CV}
}

@inproceedings{survey1,
  author={Kandwal, Siddharth and Nehra, Vibha},
booktitle={2024 14th International Conference on Cloud Computing, Data Science \& Engineering (Confluence)},
title={A Survey of Text-to-Image Diffusion Models in Generative AI}, 
  year={2024},
  volume={},
  number={},
  pages={73-78},
  doi={10.1109/Confluence60223.2024.10463372}}

@inproceedings{visualprompter, 
    title={VisualPrompter: Semantic-Aware Prompt Optimization with Visual Feedback for Text-to-Image Synthesis}, 
    author={Shiyu Wu and Mingzhen Sun and Weining Wang and Yequan Wang and Jing Liu}, 
    booktitle={International Conference on Learning Representations (ICLR)}, 
    year={2026} 
}

@misc{visiondirector,
  title        = {VisionDirector: Vision-Language Guided Closed-Loop Refinement for Generative Image Synthesis},
  author       = {Chu, Meng and Yang, Senqiao and Che, Haoxuan and Zhang, Suiyun and Zhang, Xichen and Yu, Shaozuo and Gui, Haokun and Rao, Zhefan and Tu, Dandan and Liu, Rui and Jia, Jiaya},
  year         = {2025},
  eprint       = {2512.19243},
  archivePrefix= {arXiv},
  primaryClass = {cs.CV}
}

@inproceedings{T2Icopilot,
  title={T2I-Copilot: A Training-Free Multi-Agent Text-to-Image System for Enhanced Prompt Interpretation and Interactive Generation},
  author={Chieh-Yun Chen and Min Shi and Gong Zhang and Humphrey Shi},
  booktitle={Proceedings of the IEEE/CVF International Conference on Computer Vision (ICCV)},
  year={2025}
}

@inproceedings{dsg,
author        = {Jaemin Cho and Yushi Hu and Roopal Garg and Peter Anderson and Ranjay Krishna and Jason Baldridge and Mohit Bansal and Jordi Pont-Tuset and Su Wang},
title         = {{Davidsonian Scene Graph: Improving Reliability in Fine-Grained Evaluation for Text-to-Image Generation}},
booktitle     = {ICLR},
year          = {2024}
}

@article{tifa,
title={TIFA: Accurate and Interpretable Text-to-Image Faithfulness Evaluation with Question Answering},
author={Hu, Yushi and Liu, Benlin and Kasai, Jungo and Wang, Yizhong and Ostendorf, Mari and Krishna, Ranjay and Smith,
Noah A},
journal={arXiv preprint arXiv:2303.11897},
year={2023}
}

@misc{openai2025gptimage1,
  title        = {GPT Image 1},
  author       = {{OpenAI}},
  year         = {2025},
  howpublished = {\url{https://developers.openai.com/api/docs/models/gpt-image-1}},
  note         = {OpenAI API model documentation; deprecated model}
}

@misc{titf,
      title={TIIF-Bench: How Does Your T2I Model Follow Your Instructions?}, 
      author={Xinyu Wei and Jinrui Zhang and Zeqing Wang and Hongyang Wei and Zhen Guo and Bairui Li and Lei Zhang},
      year={2026},
      eprint={2506.02161},
      archivePrefix={arXiv},
      primaryClass={cs.CV},
      url={https://arxiv.org/abs/2506.02161}, 
}

@article{oneig,
  title={OneIG-Bench: Omni-dimensional Nuanced Evaluation for Image Generation}, 
  author={Jingjing Chang and Yixiao Fang and Peng Xing and Shuhan Wu and Wei Cheng and Rui Wang and Xianfang Zeng and Gang Yu and Hai-Bao Chen},
  journal={arXiv preprint arxiv:2506.07977},
  year={2025}
}

@article{depthanything3,
  title   = {Depth Anything 3: Recovering the Visual Space from Any Views},
  author  = {Lin, Haotong and Chen, Sili and Liew, Jun Hao and Chen, Donny Y. and Li, Zhenyu and Shi, Guang and Feng, Jiashi and Kang, Bingyi},
  journal = {arXiv preprint arXiv:2511.10647},
  year    = {2025},
  url     = {https://arxiv.org/abs/2511.10647}
}

@misc{openai2026gptimage2,
  title        = {GPT Image 2},
  author       = {{OpenAI}},
  year         = {2026},
  howpublished = {\url{https://developers.openai.com/api/docs/models/gpt-image-2}},
  note         = {OpenAI API model documentation}
}

@misc{google2025gemini3proimage,
  title        = {Gemini 3 Pro Image},
  author       = {{Google DeepMind}},
  year         = {2025},
  howpublished = {\url{https://ai.google.dev/gemini-api/docs/models/gemini-3-pro-image}},
  note         = {Gemini API model documentation}
}

@misc{google2026gemini31flashimage,
  title        = {Gemini 3.1 Flash Image},
  author       = {{Google DeepMind}},
  year         = {2026},
  howpublished = {\url{https://deepmind.google/models/model-cards/gemini-3-1-flash-image/}},
  note         = {Model card}
}

@misc{google2025gemini25flashimage,
  title        = {Gemini 2.5 Flash Image},
  author       = {{Google}},
  year         = {2025},
  howpublished = {\url{https://developers.googleblog.com/en/introducing-gemini-2-5-flash-image/}},
  note         = {Google Developers Blog}
}

@misc{bytedance2026seedream5pro,
  title        = {Seedream 5.0 Pro},
  author       = {{ByteDance}},
  year         = {2026},
  howpublished = {\url{https://dreamina.capcut.com/seedream/seedream-5-0-pro}},
  note         = {Official Dreamina/CapCut model page}
}

@article{chen2025seedream4,
  title   = {Seedream 4.0: Toward Next-generation Multimodal Image Generation},
  author  = {Chen, Yunpeng and others},
  journal = {arXiv preprint arXiv:2509.20427},
  year    = {2025},
  url     = {https://arxiv.org/abs/2509.20427}
}

@misc{wu2025qwenimage,
  title         = {Qwen-Image Technical Report},
  author        = {Wu, Chenfei and Li, Jiahao and Zhou, Jingren and Lin, Junyang and Gao, Kaiyuan and Yan, Kun and Yin, Sheng-ming and Bai, Shuai and Xu, Xiao and Chen, Yilei and others},
  year          = {2025},
  eprint        = {2508.02324},
  archivePrefix = {arXiv},
  primaryClass  = {cs.CV},
  url           = {https://arxiv.org/abs/2508.02324}
}

@misc{baidu2026ernieimage,
  title        = {ERNIE-Image},
  author       = {{ERNIE-Image Team, Baidu}},
  year         = {2026},
  howpublished = {\url{https://huggingface.co/baidu/ERNIE-Image}},
  note         = {Hugging Face model card}
}

@misc{wei2025unipic2,
  title         = {Skywork UniPic 2.0: Building Kontext Model with Online RL for Unified Multimodal Model},
  author        = {Wei, Hongyang and Xu, Baixin and Liu, Hongbo and Wu, Cyrus and Liu, Jie and Peng, Yi and Wang, Peiyu and Liu, Zexiang and He, Jingwen and Xietian, Yidan and Tang, Chuanxin and Wang, Zidong and Wei, Yichen and Hu, Liang and Jiang, Boyi and Li, William and He, Ying and Liu, Yang and Song, Xuchen and Li, Eric and Zhou, Yahui},
  year          = {2025},
  eprint        = {2509.04548},
  archivePrefix = {arXiv},
  primaryClass  = {cs.CV},
  url           = {https://arxiv.org/abs/2509.04548}
}

@misc{han2024infinity,
  title         = {Infinity: Scaling Bitwise AutoRegressive Modeling for High-Resolution Image Synthesis},
  author        = {Han, Jian and Liu, Jinlai and Jiang, Yi and Yan, Bin and Zhang, Yuqi and Yuan, Zehuan and Peng, Bingyue and Liu, Xiaobing},
  year          = {2024},
  eprint        = {2412.04431},
  archivePrefix = {arXiv},
  primaryClass  = {cs.CV},
  url           = {https://arxiv.org/abs/2412.04431}
}

@article{chen2025januspro,
  title   = {Janus-Pro: Unified Multimodal Understanding and Generation with Data and Model Scaling},
  author  = {Chen, Xiaokang and Wu, Zhiyu and Liu, Xingchao and Pan, Zizheng and Liu, Wen and Xie, Zhenda and Yu, Xingkai and Ruan, Chong},
  journal = {arXiv preprint arXiv:2501.17811},
  year    = {2025},
  url     = {https://arxiv.org/abs/2501.17811}
}

@misc{flux2024,
  author       = {{Black Forest Labs}},
  title        = {FLUX},
  year         = {2024},
  howpublished = {\url{https://github.com/black-forest-labs/flux}}
}

@article{xie2024showo,
  title   = {Show-o: One Single Transformer to Unify Multimodal Understanding and Generation},
  author  = {Xie, Jinheng and Mao, Weijia and Bai, Zechen and Zhang, David Junhao and Wang, Weihao and Lin, Kevin Qinghong and Gu, Yuchao and Chen, Zhijie and Yang, Zhenheng and Shou, Mike Zheng},
  journal = {arXiv preprint arXiv:2408.12528},
  year    = {2024},
  url     = {https://arxiv.org/abs/2408.12528}
}

@misc{esser2024sd3,
  title         = {Scaling Rectified Flow Transformers for High-Resolution Image Synthesis},
  author        = {Esser, Patrick and Kulal, Sumith and Blattmann, Andreas and Entezari, Rahim and Müller, Jonas and Saini, Harry and Levi, Yam and Lorenz, Dominik and Sauer, Axel and Boesel, Frederic and Podell, Dustin and English, Zion and Lacey, Kyle and Goodwin, Alex and Marek, Yannik and Rombach, Robin},
  year          = {2024},
  eprint        = {2403.03206},
  archivePrefix = {arXiv},
  primaryClass  = {cs.CV},
  url           = {https://arxiv.org/abs/2403.03206}
}

@misc{gan,
      title={Generative Adversarial Networks}, 
      author={Ian J. Goodfellow and Jean Pouget-Abadie and Mehdi Mirza and Bing Xu and David Warde-Farley and Sherjil Ozair and Aaron Courville and Yoshua Bengio},
      year={2014},
      eprint={1406.2661},
      archivePrefix={arXiv},
      primaryClass={stat.ML},
      url={https://arxiv.org/abs/1406.2661}, 
}

@misc{rombach2022highresolutionimagesynthesislatent,
      title={High-Resolution Image Synthesis with Latent Diffusion Models}, 
      author={Robin Rombach and Andreas Blattmann and Dominik Lorenz and Patrick Esser and Björn Ommer},
      year={2022},
      eprint={2112.10752},
      archivePrefix={arXiv},
      primaryClass={cs.CV},
      url={https://arxiv.org/abs/2112.10752}, 
}

@misc{lipman2023flowmatchinggenerativemodeling,
      title={Flow Matching for Generative Modeling}, 
      author={Yaron Lipman and Ricky T. Q. Chen and Heli Ben-Hamu and Maximilian Nickel and Matt Le},
      year={2023},
      eprint={2210.02747},
      archivePrefix={arXiv},
      primaryClass={cs.LG},
      url={https://arxiv.org/abs/2210.02747}, 
}

@article{taylor1996perspective,
  title   = {Perspective in Spatial Descriptions},
  author  = {Taylor, Holly A. and Tversky, Barbara},
  journal = {Journal of Memory and Language},
  volume  = {35},
  number  = {3},
  pages   = {371--391},
  year    = {1996},
  doi     = {10.1006/jmla.1996.0021}
}

@article{tenbrink2011reference,
  title   = {Reference Frames of Space and Time in Language},
  author  = {Tenbrink, Thora},
  journal = {Journal of Pragmatics},
  volume  = {43},
  number  = {3},
  pages   = {704--722},
  year    = {2011},
  doi     = {10.1016/j.pragma.2010.06.020}
}

@article{carlsonradvansky1997reference,
  title   = {The Influence of Reference Frame Selection on Spatial Template Construction},
  author  = {Carlson-Radvansky, Laura A. and Logan, Gordon D.},
  journal = {Journal of Memory and Language},
  volume  = {37},
  number  = {3},
  pages   = {411--437},
  year    = {1997},
  doi     = {10.1006/jmla.1997.2519}
}

@misc{zhang2023texttoimage,
  title         = {Text-to-image Diffusion Models in Generative AI: A Survey},
  author        = {Zhang, Chenshuang and Zhang, Chaoning and Zhang, Mengchun and Kweon, In So},
  year          = {2023},
  eprint        = {2303.07909},
  archivePrefix = {arXiv},
  primaryClass  = {cs.CV},
  url           = {https://arxiv.org/abs/2303.07909}
}

@article{yang2023diffusionsurvey,
  title   = {Diffusion Models: A Comprehensive Survey of Methods and Applications},
  author  = {Yang, Ling and Zhang, Zhilong and Song, Yang and Hong, Shenda and Xu, Runsheng and Zhao, Yue and Zhang, Wentao and Cui, Bin and Yang, Ming-Hsuan},
  journal = {ACM Computing Surveys},
  volume  = {56},
  number  = {4},
  pages   = {1--39},
  year    = {2023},
  doi     = {10.1145/3626235}
}

@misc{cao2024controllable,
  title         = {Controllable Generation with Text-to-Image Diffusion Models: A Survey},
  author        = {Cao, Pu and Zhou, Feng and Song, Qing and Yang, Lu},
  year          = {2024},
  eprint        = {2403.04279},
  archivePrefix = {arXiv},
  primaryClass  = {cs.CV},
  url           = {https://arxiv.org/abs/2403.04279}
}

@article{liu2026diffusionrobotics,
  title   = {Diffusion Models in Robotics: A Survey},
  author  = {Liu, Xiaokang and Ma, Kevin Yuchen and Gao, Chen and Shou, Mike Zheng},
  journal = {International Journal of Computer Vision},
  year    = {2026},
  doi     = {10.1007/s11263-026-02893-1}
}

@article{ding2025worldmodels,
  title   = {Understanding World or Predicting Future? A Comprehensive Survey of World Models},
  author  = {Ding, Jingtao and Zhang, Yunke and Shang, Yu and Zhang, Yuheng and Zong, Zefang and Feng, Jie and Yuan, Yuan and Su, Hongyuan and Li, Nian and Sukiennik, Nicholas and others},
  journal = {ACM Computing Surveys},
  volume  = {58},
  number  = {3},
  pages   = {1--38},
  year    = {2025}
}

@misc{wu2025qwenimagetechnicalreport,
      title={Qwen-Image Technical Report}, 
      author={Chenfei Wu and Jiahao Li and Jingren Zhou and Junyang Lin and Kaiyuan Gao and Kun Yan and Sheng-ming Yin and Shuai Bai and Xiao Xu and Yilei Chen and Yuxiang Chen and Zecheng Tang and Zekai Zhang and Zhengyi Wang and An Yang and Bowen Yu and Chen Cheng and Dayiheng Liu and Deqing Li and Hang Zhang and Hao Meng and Hu Wei and Jingyuan Ni and Kai Chen and Kuan Cao and Liang Peng and Lin Qu and Minggang Wu and Peng Wang and Shuting Yu and Tingkun Wen and Wensen Feng and Xiaoxiao Xu and Yi Wang and Yichang Zhang and Yongqiang Zhu and Yujia Wu and Yuxuan Cai and Zenan Liu},
      year={2025},
      eprint={2508.02324},
      archivePrefix={arXiv},
      primaryClass={cs.CV},
      url={https://arxiv.org/abs/2508.02324}, 
}

@misc{carion2025sam3segmentconcepts,
      title={SAM 3: Segment Anything with Concepts},
      author={Nicolas Carion and Laura Gustafson and Yuan-Ting Hu and Shoubhik Debnath and Ronghang Hu and Didac Suris and Chaitanya Ryali and Kalyan Vasudev Alwala and Haitham Khedr and Andrew Huang and Jie Lei and Tengyu Ma and Baishan Guo and Arpit Kalla and Markus Marks and Joseph Greer and Meng Wang and Peize Sun and Roman Rädle and Triantafyllos Afouras and Effrosyni Mavroudi and Katherine Xu and Tsung-Han Wu and Yu Zhou and Liliane Momeni and Rishi Hazra and Shuangrui Ding and Sagar Vaze and Francois Porcher and Feng Li and Siyuan Li and Aishwarya Kamath and Ho Kei Cheng and Piotr Dollár and Nikhila Ravi and Kate Saenko and Pengchuan Zhang and Christoph Feichtenhofer},
      year={2025},
      eprint={2511.16719},
      archivePrefix={arXiv},
      primaryClass={cs.CV},
      url={https://arxiv.org/abs/2511.16719},
}

\end{document}